\documentclass[runningheads]{llncs}

\usepackage[T1]{fontenc}
\usepackage{graphicx}
\usepackage[utf8]{inputenc} 
\usepackage[T1]{fontenc}    
\usepackage{hyperref}       
\usepackage{url}            
\usepackage{booktabs}       
\usepackage{amsfonts}       
\usepackage{nicefrac}       
\usepackage{microtype}      
\usepackage{xcolor}         
\usepackage{wrapfig}

\usepackage{caption}
\usepackage{subcaption}

\usepackage{amsmath}
\usepackage{amsfonts}
\usepackage{amssymb}
\usepackage{xfrac}

\usepackage{pdflscape}
\usepackage{pdfpages}
\usepackage{longtable}

\usepackage{algorithm}
\usepackage{algpseudocode}

\usepackage{paralist}
\usepackage[shortlabels]{enumitem}

\newcommand{\N}{\mathbb{N}}
\newcommand{\R}{\mathbb{R}}
\renewcommand{\P}{\mathbb{P}}
\newcommand{\D}{\mathcal{D}}
\renewcommand{\d}{\textnormal{d}}
\newcommand{\I}{\mathbb{I}}
\newcommand{\E}{\mathbb{E}}
\renewcommand{\d}{\textnormal{d}}

\renewcommand{\H}{\mathcal{H}}
\newcommand{\1}{\mathbf{1}}

\newcommand{\id}{\textnormal{id}}
\newcommand{\MSE}{\textnormal{MSE}}
\newcommand{\Var}{\textnormal{Var}}
\newcommand{\lMSE}{\textnormal{l}\MSE}

\newcommand{\indp}{\perp\!\!\!\perp}

\newcommand{\A}{\mathcal{A}}

\newcommand{\X}{\mathcal{X}}

\newcommand{\T}{\mathcal{T}}

\newcommand{\F}{\mathcal{F}}

\usepackage{todonotes}

\newcommand{\OOproof}[1]{\begin{proof}#1\end{proof}}

\usepackage{tikz}
\usetikzlibrary{positioning,shadows,shapes.symbols,arrows.meta}
\usepackage{amssymb}
\usepackage{pifont}
\newcommand{\cmark}{\ding{51}}%
\newcommand{\xmark}{\ding{55}}%

\usepackage{apxproof}
\usepackage{cleveref}

\newcommand\blfootnote[1]{%
  \begingroup
  \renewcommand\thefootnote{}\footnote{#1}%
  \addtocounter{footnote}{-1}%
  \endgroup
}

\begin{document}

\title{A Remark on Concept Drift for Dependent Data\thanks{Funding in the frame of the ERC Synergy Grant ``Water-Futures'' No. 951424 is gratefully acknowledged.}}
\author{Fabian Hinder\orcidID{0000-0002-1199-4085}${}^{,\dagger}$\and 
Valerie Vaquet\orcidID{0000-0001-7659-857X}\and 
Barbara Hammer\orcidID{0000-0002-0935-5591}}
\authorrunning{F. Hinder et al.}
\institute{CITEC, Bielefeld University, Bielefeld, Germany \\
\email{\{fhinder,vvaquet,bhammer\}@techfak.uni-bielefeld.de}}
\maketitle              
\blfootnote{\!\!\!\!${}^\dagger$ Corresponding Author}
\begin{abstract}
Concept drift, i.e., the change of the data generating distribution, can render machine learning models inaccurate. 
Several works address the phenomenon of concept drift in the streaming context usually assuming that consecutive data points are independent of each other. To generalize to dependent data, many authors link the notion of concept drift to time series. 
In this work, we show that
the temporal dependencies are strongly influencing the sampling process. Thus, the used definitions need major modifications. 
In particular, we show that the notion of stationarity is not suited for this setup and discuss alternatives. We demonstrate that these alternative formal notions describe the observable learning behavior in numerical experiments.
\end{abstract}

\section{Introduction}
\label{sec:intro}

The world surrounding us is subject to constant change, which affects the increasing amount of data collected over time. 
As those changes, referred to as \emph{concept drift} or drift for shorthand \cite{gama2014survey}, constitute a major issue in many applications, considerable research is focusing on this phenomenon. While in some applications, the goal is keeping an accurate model in the presence of drifts (\emph{stream or online leaning})~\cite{DBLP:journals/cim/DitzlerRAP15}, in \emph{monitoring setups} the goal is detecting and gathering information about the drift to inform appropriate reactions~\cite{OneOrTwoThings}. 

Frequently, the general assumption is that independent observations are collected over time even if they arrive in direct succession \cite{DBLP:journals/cim/DitzlerRAP15,OneOrTwoThings}. However, in many real-world scenarios, temporal dependencies in the data have to be expected. This is for example the case when considering \emph{time series} 
\cite{lim2021time,aminikhanghahi2017survey,shalev2014understanding}. In this setting, the absence of change across multiple time series is described by the concept of \emph{stationarity}. 
This notion is relevant if the goal is to obtain knowledge about the time series-generating process. This does not align with the goal of machine learning and monitoring, where we want to keep an up-to-date model based on our observations or detect anomalous behavior therein. In particular, we typically observe only one time series rather than multiple instantiations of time series. In this work, we  analyze the connection of drift and stationarity in settings relevant to machine learning applications. We propose the notion of \emph{consistency} when dealing with streaming data which stems from one time series.  Moreover, we demonstrate in experiments, that this formalization captures the intuitive notion of drift for possibly dependent data streams.

This paper is organized as follows: In \cref{sec:problem-setup} we recap the concept of drift and stochastic processes used later on. We categorize the possible setups and analyze them with respect to their convergence properties (\cref{sec:taxonomy}). In \cref{sec:consistency}, we provide examples showcasing that stationarity and drift do not align in settings of interest for machine learning (\cref{sec:stationarity}). Thus, we propose the notion of consistency when dealing with a data stream possibly subject to time dependencies (\cref{sec:def-consistency}). We derive a first method checking for this concept (\cref{sec:method}). Finally, we confirm the theoretical arguments with numerical experiments (\cref{sec:exp}) and conclude the work (\cref{sec:concl}).

\section{Problem Setup\label{sec:problem-setup}}
In this work, we consider the problem of stream learning, i.e., at every given time point $\tau$ the data $(X_i,T_i)$ observed until $\tau$, i.e., $T_i \leq \tau$, is available. The observations $X_i$ can be independent of each other, dependent on the past (time series), or the time point $T_i$ of observation (drift). Depending on the specific setup, learning faces different problems, e.g., model adaption. In the following, we  first give a precise formal definition of the notion of drift (of independent data) and stochastic processes/time series. We then discuss the challenges of defining drift for dependent data.

\subsection{A Probability Theoretical Framework for Concept Drift}
\label{sec:IntroDrift}

In the classical setup, machine learning assumes a time-invariant distribution $\D$ on the data space $\X$ and data points as i.i.d. samples drawn from it.
This assumption is violated in many real-world applications, in particular, when learning on data streams.
As a consequence, machine learning models can become inaccurate over time. As a first step to tackle this issue, we incorporate time into our considerations. 
To do so, let $\T$ be an index set representing time and a (potentially different) distribution $\D_t$ for every $t \in \T$. Drift refers to the phenomenon that those distributions differ for different points in time~\cite{gama2014survey}. It is possible to extend this setup to a general statistical interdependence of data and time via a distribution $\D$ on $\T \times \X$ which decomposes into a distribution $P_T$ on $\T$ and the conditional distributions $\D_t$ on $\X$  \cite{DAWIDD,OneOrTwoThings}. One of the key findings of \cite{DAWIDD} is a unique characterization of the presence of drift by the property of statistical dependency of time $T$ and data $X$. 

\begin{definition}
\label{def:drift}
Let $\X, \T$ be countably generated measurable spaces.
Let $(\D_t,P_T)$ be a \emph{drift process}~\cite{DAWIDD} during $\T$ on $\X$, i.e.,\ a distribution $P_T$ on $\T$ and Markov kernel $\D_t$ from $\T$ to $\X$. 
A \emph{time window} $W \subset \T$ is a $P_T$ non-null set. A \emph{sample (of size $N$)} is an $\T \times \X$ $N$-tuple $S = ((T_1,X_1),\dots,(T_N,X_N))$ with $T_i \sim P_T, X_i \mid [T_i = t] \sim \D_t$ i.i.d. 
We say that $\D_t$ has \emph{drift} if the probability of observing two different distributions in one sample is larger 0, i.e., $\P_{T,S \sim P_T} [\D_T \neq \D_S] > 0$.
\end{definition}

Considerable work focuses on independent data streams with drift: Many authors consider the model accuracy as a proxy for drift and update the model if a decline in the accuracy is observed~\cite{DBLP:journals/cim/DitzlerRAP15}. As the connection between model loss and drift in the data generating distribution is rather lose \cite{hinder2023hardness,ida2023}, relying on distribution-based drift detection schemes instead of model loss-based schemes enables reliable monitoring~\cite{OneOrTwoThings}. Assuming independent data points $X_i$, allows applying tools from classical statistics or learning theory~\cite{DAWIDD,ida2023}. In contrast, given temporal dependencies in the observed data, these strategies cannot be used successfully. 

\subsection{Stochastic Processes}
Instead of drift processes, dependent data streams or time series can be modeled by stochastic processes~\cite{borodin2017stochastic}:

\begin{definition}
    \label{def:sp}
    Let $(\Omega,\A,\P)$ be a probability space, $(\X,\Sigma_\X)$ a measurable space, and $\T$ an index-set. A \emph{$\X$-valued stochastic process} over $\T$ is a map $X_\bullet : \T \times \Omega \to \X$ such that $X_t : \Omega \to \X$ is a $\X$-valued random variable for all $t \in  \T$. We call the maps $t \mapsto X_t(\omega)$ the \emph{paths} of $X_t$. For a finite, enumerated subset $\T_0 \subset \T$ we refer to a realization of $X_t$ restricted to $\T_0$ as a \emph{time series}, i.e., $(x_i)_{i = 1}^{N} = (X_{t_i}(\omega))_{i=1}^N$ where $\T_0 = \{t_1,\dots,t_N\}$.
    The process is called \emph{stationary} iff the distribution of $X_t$ is time invariant, i.e., $\P[X_t \in A] = \P[X_s \in A]$ for all $s,t \in \T, A \subset \X$. For $\T \subset \R$ we call $t \in \T$ a \emph{change point} iff the distribution changes at $t$.
    Assuming $(\T,\Sigma_\T)$ is a measurable space, we call $X_t$ \emph{measurable} iff $X_\bullet$ is measurable with respect to $\A \otimes \Sigma_\T$. 
\end{definition}

A measurable stochastic process is a random variable that takes values in the space of measurable functions from $\T$ to $\X$. 
In particular, measurability is required if we want to integrate along the path as done in ergodic processes. 

The main difference between stochastic and drift processes is that a stochastic process has one value for every point in time whereas a drift process yields a distribution. For example: Measuring the temperature of an object over time is a stochastic process because we read exactly one value. In contrast, a stream of ballots should be considered a drift process because the distribution is more interesting than a single vote.

\begin{figure}[t]
    \centering
    \begin{tikzpicture}[node distance=0.5cm]
    
        \tikzstyle{every node}=[font=\scriptsize]
        \node (datastream) at (0,0) {data stream};
        \node [below left = 0.5cm and 1cm of datastream] (independentpoints) {\begin{tabular}{c}independent\\data points\\\cite{gama2014survey,DBLP:journals/cim/DitzlerRAP15,OneOrTwoThings}\end{tabular}};
        \node [below right = 0.5cm and 1cm of datastream] (dependentpoints) {\begin{tabular}{c}dependent\\data points\end{tabular}};
        \draw[-{Latex[length=2mm]}] (datastream) -- (independentpoints.north);
        \draw[-{Latex[length=2mm]}] (datastream) -- (dependentpoints.north);
        
        \node [below left = 0.5cm and 1cm   of dependentpoints] (multistream) {\begin{tabular}{c}multiple\\independent\\streams\end{tabular}};
        \node [below right = 0.5cm and 1cm  of dependentpoints] (onestream) {\begin{tabular}{c}one\\stream\end{tabular}};
        \draw[-{Latex[length=2mm]}] (dependentpoints) -- (multistream.north);
        \draw[-{Latex[length=2mm]}] (dependentpoints) -- (onestream.north);

        \node [below left = of onestream] (ergodic) {\begin{tabular}{c}ergodic\end{tabular}};
        \node [below right = of onestream] (notergodic) {\begin{tabular}{c}not\\ergodic\end{tabular}};
        \draw[-{Latex[length=2mm]}] (onestream) -- (ergodic.north);
        \draw[-{Latex[length=2mm]}] (onestream) -- (notergodic.north);

        \node [below left = of ergodic] (mixing) {\begin{tabular}{c}mixing\\\cite{hanneke2019statistical,agarwal2012generalization}\end{tabular}};
        \node [below right = of ergodic] (notmixing) {\begin{tabular}{c}not\\mixing\\\cite{adams2010uniform}\end{tabular}};
        \draw[-{Latex[length=2mm]}] (ergodic) -- (mixing.north);
        \draw[-{Latex[length=2mm]}] (ergodic) -- (notmixing.north);

        \node [below left= 4cm and -0.5cm of independentpoints] (reasonable) {\small$=$};
        \node [right = 11cm of reasonable] (nonsense) {\small$\neq$};
        \node [above right = 0cm and 4.4cm of reasonable]{drift (Def. \ref{def:drift}) = stationarity (Def. \ref{def:sp})};
        \draw[{Latex[length=2mm]}-{Latex[length=2mm]}] (reasonable.north west) -- (nonsense.north east);
        \node [right = 2cm of reasonable] {\cmark};
        \node [right = 6.5cm of reasonable] {\textbf{?}};
        \node [left = of nonsense] {\xmark};

        \node[above right = 0.1cm and 0.3cm of multistream] (upperbar1){};
        \node[below = 4cm of upperbar1] (lowerbar1){};
        \draw[dashed] (upperbar1) -- (lowerbar1);

        \node[below right = 1cm and 4.4cm of upperbar1] (upperbar2){};
        \node[right = 4.4cm of lowerbar1] (lowerbar2){};
        \draw[dashed] (upperbar2) -- (lowerbar2);

        \draw[-{Latex[length=2mm]}, gray] (multistream) edge [out=180,in=0] node[midway,left] {\begin{tabular}{c}
             subsampling\\
             (Thm. \ref{thm:sampling}) 
        \end{tabular}
        } (independentpoints);
    \end{tikzpicture}
    \caption{Taxonomy of change detection setups in data streams}
    \label{fig:taxonomy}
\end{figure}
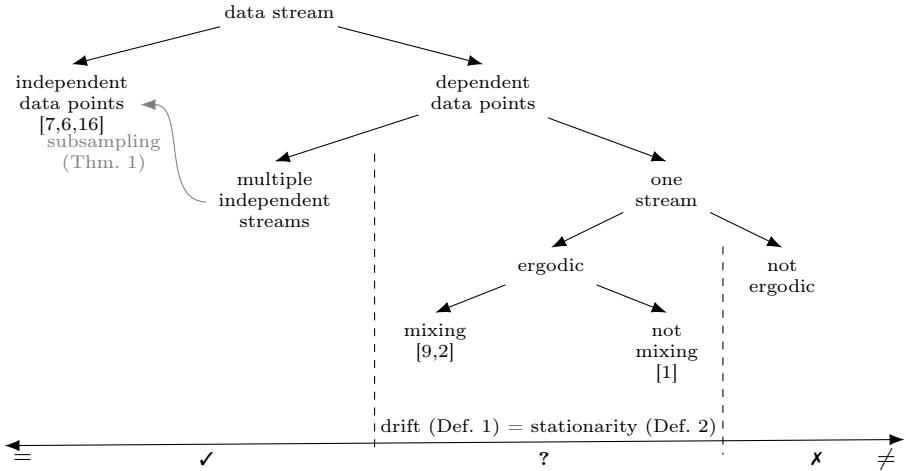
\subsection{A taxonomy of change detection in data streams\label{sec:taxonomy}}
Different setups can be considered in the context of concept drift and dependent data. 
Intuitively, the absence of stationarity is the same as the presence of drift. We  investigate this relationship in more detail: We  first propose a categorization of the possible setups as visualized in \cref{fig:taxonomy}. In the remainder of this section, we  describe the different setups.

\paragraph{Independent data points}
The classical setup of concept drift states that we have several independent data points $X_1,\dots,X_N,\dots$ that are sampled from potentially different distributions. This can be investigated for general machine learning~\cite{gama2014survey,DBLP:journals/cim/DitzlerRAP15,ida2023} or for drift detection~\cite{DAWIDD,MMD,KCpD,KFRD,OneOrTwoThings}.
As we can characterize a drift process solely by the values of all functions $f : \X \to \R$ integrated over all time windows $W$~\cite{DAWIDD,hinder2023hardness} 
we can test for drift as follows:
instead of only tracking the data points we additionally consider the time of observation, $T_1,\dots,T_N,\dots$. Using this information we can compute the value
\begin{align}
    \frac{\sum_{i = 1}^N f(X_i(\omega)) \1[T_i(\omega) \in W]}{\sum_{i = 1}^N \1[T_i(\omega) \in W]} \xrightarrow{N \to \infty} 
    &\iint f(x) \d \D_t(x) \d P_T(t \mid W) & \text{ $\P$-a.s.} \label{eq:drift_double_integral}
    \\&=\!\!= \E[f(X)\mid T \in W], \nonumber
\end{align}
where $\omega$ is the random-event. 
In particular, the limit always applies, and the value on the right-hand side does not depend on $\omega$ only on the drift process $(\D_t,P_T)$ (and $f$ and $W$). 
This justifies defining drift as $\D_{T_1} \neq \D_{T_2}$ being observed with a probability larger than zero. Thus, in an independent data stream, the presence drift is the same as the absence of stationarity (up to a $P_T$ null set).

\paragraph{Multiple (independent) streams} 
In real-world applications, data points frequently are not independent. Instead of observing an independent data stream, we observe one or multiple stochastic processes. 
Let us consider the case of multiple ones first: Let $X_t^{(1)},\dots,X_t^{(M)},\dots$ be sampled independently from the same distribution and each modeling the temporal dependency of consecutive observations. 
This setup can be reduced to the independent data streams as above:
\begin{theorem}
\label{thm:sampling}
    Let $\T$ and $\X$ be countably generated, e.g., $\T = [0,1], \N, \R,\; \X = \R^d$.
    Let $X_t^{(1)},\dots,X_t^{(M)},\dots$ i.i.d. measurable stochastic processes and $P_T$ any probability measure on $\T$. Let $T_1,\dots,T_M,\dots \sim P_T$ be i.i.d. and denote by $X_i = X_{T_i}^{(i)}$. Then all $(T_i,X_i)$ are i.i.d. and $X_i \mid [T_i = t] \sim \P_{X_t}$ for all $t \in \T$, i.e., they form a sample of the drift process $(\P_{X_t},P_T)$. 
    
    Furthermore, $(\P_{X_t},P_T)$ has drift (in the sense of \cref{def:drift}) if and only if there exists a $P_T$ null-set $N$ such that $X_t$ restricted onto $\T \setminus N$ is stationary. 
\end{theorem}
\OOproof{
As $T : \Omega \to \T$ is measurable, so is the graph map $\Gamma_T : \Omega \to \Omega \times \T, \omega \to (\omega,T(\omega))$ as for $A \in \A, B \in \Sigma_\T$ we have $\Gamma_T^{-1}(A \times B) = A \cap T^{-1}(B)$. And it suffices to check this on a generator. 
Applying this map twice, we have 
\begin{align*}
\Omega \xrightarrow{\Gamma_T} \Omega \times \T \xrightarrow{(\Gamma_T) \times \id_\T} (\Omega \times \T) \times \T &\xrightarrow{X_\bullet \times \id_\T} \X \times \T, \\ \omega \mapsto (\omega,T(\omega)) \mapsto (\omega,T(\omega),T(\omega)) &\mapsto (X_{T(\omega)}(\omega),T(\omega))
\end{align*}
is measurable. For $A \subset \X, W \subset \T$ measurable we have
\begin{align*}
    \P[ X_T \in A , T \in W ] &= {\E[ \I_A(X_T)\I_W(T) ]}
    \\&= {\E\left[ \E\left[ \I_A(X_T)\I_W(T) \mid T \right] \right]}
    \\&= {\E\left[ \E\left[ \I_A(X_T) \mid T \right] \I_W(T) \right]}
    \\&\overset{RN}{=} {\int_W \E\left[ \I_A(X_T) \mid T=t \right] \d P_T(t)}
    \\&\overset{T \indp X_t}{=} {\int_W  \E[\I_A(X_t)] \d P_T(t)}
    \\&= {\int_W  \P_{X_t}(A) \d P_T(t)},
\end{align*}
where $\E[\cdot\mid T=t]$ denotes the conditional expectation function given $T$ in the $L^2$ sense. Thus, $X_T \mid [T = t] \sim \P_{X_t}$.

Notice that $X_T$ is a $\X$-valued random variable. Let $J \subset \N$ be finite and w.l.o.g. $J = \{1,\dots,n\}$. Let $W_1,\dots,W_n \subset \T$ time windows, $A_1,\dots,A_n \subset \X$. Consider
\begin{align*}
    &\P[(X^{(1)}_{T_1},T_1) \in A_1\times W_1 ,\dots, (X^{(n)}_{T_n},T_n) \in A_n\times W_n]
     \\&= \E\left[\prod_{i = 1}^n \I_{A_i}(X^{(i)}_{T_i})\I_{W_i}(T_i) \right]
    \\&= \E\left[\E\left[\left.\prod_{i = 1}^n \I_{A_i}(X^{(i)}_{T_i}) \right| T_1,\dots,T_n \right] \prod_{i = 1}^n \I_{W_i}(T_i) \right]
    \\&\overset{X_t^{(i)} \text{ indep.}}{=} \E\left[\prod_{i = 1}^n  \E\left[\I_{A_i}(X^{(i)}_{T_i}) \mid T_1,\dots,T_n \right] \I_{W_i}(T_i) \right]
    \\&= \E\left[\prod_{i = 1}^n  \E\left[\I_{A_i}(X^{(i)}_{T_i}) \mid T_i \right] \I_{W_i}(T_i) \right]
    \\&= \prod_{i = 1}^n  \E\left[\E\left[\I_{A_i}(X^{(i)}_{T_i}) \mid T_i \right] \I_{W_i}(T_i) \right]
    \\&= \prod_{i = 1}^n  \P[X^{(i)}_{T_i} \in A_i, T_i \in W_i]
\end{align*}
Thus, $(X_i,T_i)$ are indeed i.i.d. with $X_i \mid [T_i = t] \sim \P_{X_t}$.

If $X_t$ is stationary, then $\P_{X_t} = \P_{X_s}$ for all $t \in \T$ thus $\P_{X_t} \neq \P_{X_s}$ is a $P_T^2$ null-set so there is no drift. Conversely, assuming there is no drift then
using the ideas of \cite{DAWIDD} there exists $t_0 \in \T$ such that $\P_{X_t} = \P_{X_{t_0}}$ for $P_T$-a.s. all $t$ and thus $\P_{X_t} = \P_{X_{t_0}} = \P_{X_s}$ for all $s,t \in \T\setminus N$ for some $P_T$-null set $N \subset \T$.
}

Thus, we  consider the case of a single dependent path which is the standard setting for time series analysis~\cite{truong2020selective,aminikhanghahi2017survey} in the following.

\begin{figure}[t]
    \centering
    \begin{tikzpicture}[node distance=1.7cm]
        \tikzstyle{every node}=[font=\scriptsize]
        \node[rectangle,draw=black,minimum height=3cm] (b)  at (0,0){
        \begin{tabular}{c}
             Time Average\\
             $\frac{1}{\tau}\!\!\int^{\tau}_0 \!f(X_t(\omega)) dt$ 
        \end{tabular} };

        \node[rectangle,draw=black,minimum height=3cm, right = of b] (c){
        \begin{tabular}{c}
            Space Average \\
             $\E[f(X) \mid T \leq \tau]$
        \end{tabular} };

        \node[above left = -0.7cm and 1.7cm of b, minimum width=2.2cm] (independent){\begin{tabular}{c}independent\\data points\end{tabular}};
        \node[below = 0.1cm of independent, minimum width=2.2cm] (mixing){\begin{tabular}{c}mixing\end{tabular}};
        \node[below = 0.1cm of mixing, minimum width=2.2cm] (notmixing){\begin{tabular}{c}not mixing\end{tabular}};
        \node[below = 0.1cm of notmixing, minimum width=2.2cm] (notergodic){\begin{tabular}{c}not ergodic\end{tabular}};
        \node[below = 0.1cm of notergodic, minimum width=2.2cm] (notmeasurable){\begin{tabular}{c}not measurable\end{tabular}};

        \node[right = 1.7cm of independent] (independenthook) {};
        \draw[-{Latex[length=3mm]}] (independent) edge  node[midway,above] {$N\rightarrow\infty$} (independenthook);
        \node[right = 1.7cm of mixing] (mixinghook) {};
        \draw[-{Latex[length=3mm]}] (mixing) edge  node[midway,above] {$N\rightarrow\infty$} (mixinghook);
        \node[right = 1.7cm of notmixing] (notmixinghook) {};
        \draw[-{Latex[length=3mm]}] (notmixing) edge  node[midway,above] {$N\rightarrow\infty$} (notmixinghook);
        \node[right = 1.7cm of notergodic] (notergodichook) {};
        \draw[-{Latex[length=3mm]}] (notergodic) edge  node[midway,above] {$N\rightarrow\infty$} (notergodichook);
        \node[right = 1.7cm of notmeasurable] (notmeasurablehook) {};
        \node[left = 0.7cm of notmeasurablehook] {\xmark};

        \node[right = 1.8cm of independenthook] (independenthook2) {};
        \node[right = 1.7cm of independenthook2] (independenthook3) {};
        \draw[-{Latex[length=3mm]}] (independenthook2) edge  node[midway,above] {$\forall \tau \!\!: \: =\!=$} (independenthook3);
        
        \node[right = 1.8cm of mixinghook] (mixinghook2) {};
        \node[right = 1.7cm of mixinghook2] (mixinghook3) {};
        \draw[-{Latex[length=3mm]}] (mixinghook2) edge  node[midway,above] {$\tau\rightarrow\infty$} node[midway,below] {uniform} (mixinghook3);

        \node[right = 1.8cm of notmixinghook] (notmixinghook2) {};
        \node[right = 1.7cm of notmixinghook2] (notmixinghook3) {};
        \draw[-{Latex[length=3mm]}] (notmixinghook2) edge  node[midway,above] {$\tau\rightarrow\infty$} node[midway,below] {non-uniform} (notmixinghook3);

        \node[right = 1.8cm of notergodichook] (notergodichook2) {};
        \node[right = 1.7cm of notergodichook2] (notergodichook3) {};
        \node[left = 0.7cm of notergodichook3] {\xmark};
        
        \node[right = 1.8cm of notmeasurablehook] (notmeasurablehook2) {};
        \node[right = 1.7cm of notmeasurablehook2] (notmeasurablehook3) {};
        \node[left = 0.7cm of notmeasurablehook3] {\xmark};
    \end{tikzpicture}
    \caption{Two stages of converges for different types of streams. Left to right: Empirical mean, path integral / time-average, mean of the underlying distribution.}
    \label{fig:process-convergence-chart}
\end{figure}
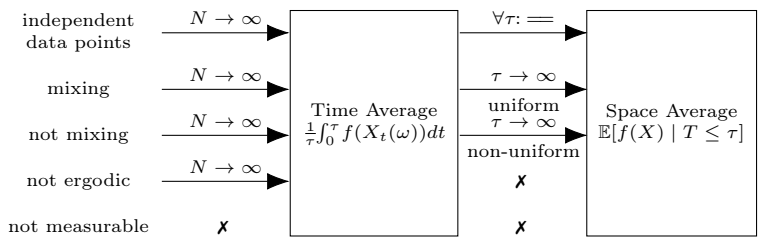

\paragraph{Ergodicity and mixing} Considering only one dependent process we first encounter a phenomenon that we refer to as the \emph{uncertainty principle of drift learning}. To understand this assume we want to estimate the value of $\E[f(X)\mid T = t]$. As the probability of making an observation exactly at time $t$ is zero, we replace it with a small time window $W$. In the independent setup, we simply sample more and more data points and gain better and better estimation (see \cref{eq:drift_double_integral}) and then successively shrink $W$ towards $t$. However, in the single-path setup, we can only sample one path at several time points. As a consequence, the inner integral on the right-hand side of \cref{eq:drift_double_integral} becomes a \emph{time-average}:
\begin{align*}
    \frac{\sum_{i = 1}^{N} f(X_{T_i(\omega)}(\omega)) \1[T_i(\omega) \in W]}{\sum_{i = 1}^{N} \1[T_i(\omega) \in W]} \xrightarrow{N \to \infty} \int f(X_t(\omega)) \d P_T(t \mid W).
\end{align*}
As can be seen, the right-hand side now does depend on $\omega$ and does not necessarily converge to the mean of the underlying distribution during $W$ also known as \emph{space average}, i.e., $\iint f(x) \d \D_t(x) \d P_T(t \mid W)$.
Intuitively speaking, dependency limits the amount of local information and due to drift, we cannot obtain more information by considering larger time windows. 
The special cases in which the convergence holds are summarized in \cref{fig:process-convergence-chart} and discussed in the following.

Processes for which such convergence holds are referred to as \emph{ergodic} processes~\cite{borodin2017stochastic}. They allow precise estimations and learning under the assumption of stationarity~\cite{adams2010uniform}. Yet, they have arbitrarily slow convergence~\cite{krengel1978speed}. 
A common property used to counteract this issue is offered by \emph{mixing processes} that impose a rate on becoming independent. This leads to uniform convergence similar to the independent case and has been used by several authors to prove statements on learnability in the stationary~\cite{yu1994rates,kontorovich2007measure} and 
non-stationary setup~\cite{hanneke2019statistical,agarwal2012generalization}. 

Yet, both ergodic and mixing processes usually require infinite temporal horizons to achieve convergence. As we are usually concerned with limited time the results might be interesting from a theoretical point of view but have little practical relevance. Also, due to the 
uncertainty principle we introduced before most make additional requirements that limit the non-stationarity in one way or another. In particular, its relation to drift detection in the monitoring setup is not clear. 

\paragraph*{Not-ergodic processes} Ergodicity usually describes that the time average equals the space average for infinitely large time windows. We will call processes for which there are either no time windows for which equality holds or for which this cannot be predicted \emph{not-ergodic}. 

It is obvious that stationarity is not a reasonable tool to discuss not-ergodic processes as there is no real connection to the underlying distributions. This implies that there cannot be any formal guarantees regarding generalization properties in the usual sense. 

\paragraph{Research questions} Based on the presented analysis we notice that there do exist theoretical results on detecting drift in dependent data but their relevance for practical setups is questionable. This leads us to the following questions that we address in the following:
\begin{enumerate}
    \item Is non-stationarity a suitable definition for the notion of concept drift for dependent data? \label{q:stat-good}
    \item Does there exist an alternative notion that is more suitable for practical application (this includes testability on finite temporal horizons)? \label{q:algo}
\end{enumerate}

\section{Consistency Property\label{sec:consistency}}
In this section, we investigate the suitability of stationarity to model the equivalent of drift as used for independent data streams on time series. 
First, we provide (counter-)examples that show that this is not the case.
Based thereon, we derive the concept of temporal consistency to formalize drift for time series. Finally, we derive a method to test consistency.

\subsection{Drift is not Non-Stationarity\label{sec:stationarity}}
\begin{wrapfigure}[11]{R}{0.45\textwidth}
  \begin{center}
    \vspace{-3.5em}
    \includegraphics[width=0.45\textwidth]{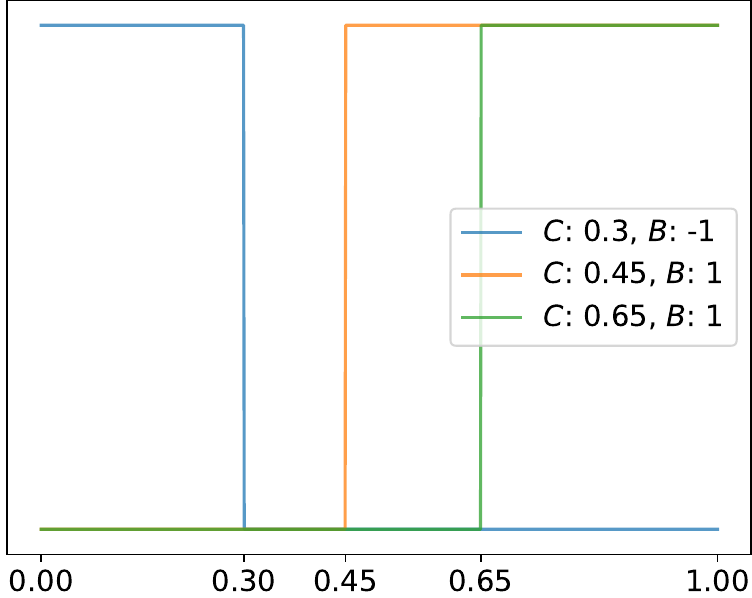}
  \end{center}
  \vspace{-1.5em}
  \caption{Sample paths from \cref{emp:jump}.}
  \label{fig:jump}
\end{wrapfigure}

That the presence of drift is not the same as the absence of stationarity for dependent data streams can be shown by the following, simple counterintuitive examples:
\begin{example}
\label{emp:jump}
    Let $\X = \{-1,1\}, \T = [0,1]$ and $B$ and $C$ independent random variables with $\P[B = -1] = p, \P[B = 1] = 1-p$ for $p \in [0,1]$, and $C \sim \mathcal{U}(\T)$. Consider $X_t = (1 - 2 \cdot \1[C > t]) \cdot B$ the process whose paths take values in $\{-1,1\}$ and switch sign exactly once (see \cref{fig:jump}). Then we have $\P[X_t = -1] = t(1-p)+(1-t)p$ and therefore $X_t$ is stationary if and only if $p = \frac{1}{2}$ and otherwise every $t \in \T$ is a change point. Yet, every path has exactly one jump at time $C$.
\end{example}
\OOproof{
Notice that $\P[X_t = B \mid t \leq C] = 1$ and $\P[X_t = B \mid t > C] = 0$ for all $t$ by definition. Hence, $\P[X_t = B] = 
\P[X_t = B \mid t \leq C]\P[t \leq C] + \P[X_t = B \mid t > C]\P[t > C]
= 1\cdot(t) + 0\cdot(1-t) = t$ and $\P[X_t \neq B] = 1-t$ for all $t$. Therefore, $\P[X_t = 1] = \P[X_t = 1 \mid B = 1]\P[B=1] + \P[X_t = 1 \mid B=-1]\P[B=-1] = t(1-p) + (1-t)p$ for all $t$. By considering $t=0,1$ we see that stationarity holds if and only if $p = 1-p \Leftrightarrow p = \frac{1}{2}$
}
\begin{example}
    \label{emp:teach}
    Consider a teacher student scenario: Let $\X = \R^d \times \{0,1\}, \; \T = \mathbb{Z}$. Let $X_t \sim \mathcal{U}([-1,1]^d), X_t \indp X_s \forall t \neq s$ and $y_t = \1[w_t^\top X_t > 0]$ where $w_t$ is the teacher weight vector which is independently resampled every $N > 1$ time steps, i.e., $w_{iN+j} = w_{iN}\forall i \in \mathbb{Z}, j = 0,\dots,N-1,\;w_{iN} \sim \mathcal{N}(0,1), \; w_t \indp w_s \forall |t-s| > N$. Then $(X_t,y_t)$ is a labeled data stream that is $\psi$-mixing and stationary. Yet, every $N$ time steps the teacher is changed in a completely random way.
\end{example}
\OOproof{
Since $y_t = \1[w_t^\top X_t > 0]$ we have 
\begin{align*}
&(X_t,y_t) \indp (X_s,y_s) \\&\Leftrightarrow (X_t,w_t) \indp (X_s,w_s) \\&\Leftrightarrow (w_t \indp w_s \mid X_s,X_t) \wedge (X_t \indp w_s \mid X_s) \wedge (w_t \indp X_s \mid X_t) \wedge (X_t \indp X_s) \\&\Leftrightarrow w_t \indp w_s \Leftrightarrow |s-t| > N.
\end{align*}
Thus, for all $s \geq t+N$ and $A,B \subset \X$ we have $\P[(X_t,y_t) \in A, (X_s,y_s) \in B] = \P[(X_t,y_t) \in A]\P[(X_s,y_s) \in B]$ so the process is $\psi$-mixing. 
Now, since $\P[X_t \in A, y_t = 1] = \iint \1[w^\top x > 0]\I_A(x) \d \mathcal{N}(w;0,1) \d \mathcal{U}(x ; [-1,1]^d)$ and the right-hand side does not depend on $t$ the process is stationary.
}

Both examples show a common theme: If we consider one path only then there are obvious changes, but if we average over several paths then those changes even out (at least for $p=\sfrac{1}{2}$). 
Thus, we can conclude that in contrast to stationarity, drift is a path-wise notion.
A na\"ive way to solve this problem is to apply the definition of stationarity path-wise, i.e., $\P[X_S = X_T] = 1$. However, such processes are essentially constant. Therefore the notion is far too restrictive to be of interest: 
\begin{lemma}
    Let $\T$ and $\X$ be countably generated with $\X$ having a measurable diagonal, i.e., $\T = [0,1], \N, \R,\; \X = \R^d$.
    Let $X_t$ be a measurable stochastic process and $P_T$ a measure on $\T$ with $S,T \sim P_T$ independent of each other and $X_t$. It holds $X_S = X_T$ $\P$-a.s., i.e., $\P \times P_T^2(\{(\omega,s,t) \mid X_t(\omega) = X_s(\omega) \}) = 1$, if and only if $X_t$ is constant for $P_T$-a.e. $t$, i.e., there exists a random variable $C$ such that $\P\times P_T(\{(\omega,t) \mid X_t(\omega) = C(\omega)\}) = 1$.
\end{lemma}
\OOproof{
\newcommand{\PPP}{\P\times P_T^2}
\newcommand{\PP}{\P \times P_T}
Due to the measurability of $\Delta_\X$ the equality sets are measurable.

Assuming $C$ exists, then we have 
\begin{align*}
  &     \PPP(\{ (\omega,s,t) \mid X_t(\omega) = X_s(\omega) \})
\\&\geq \PPP(\{ (\omega,s,t) \mid X_t(\omega) = C(\omega) , X_s(\omega) = C(\omega) \})
\\&   = \PPP(\{ (\omega,s,t) \mid X_t(\omega) = C(\omega) , s \in \T \} \cap \{ (\omega,s,t) \mid t \in \T , X_s(\omega) = C(\omega) \}) 
\\&\overset{!}{\geq} 1-2\PP(\{ (\omega,t) \mid X_t(\omega) \neq C(\omega)\}) 
\\&= 1-2 \cdot 0 = 1,
\end{align*}
where $!$ holds because $\mu(A \cap B) = 1-\mu( \;(A \cap B)^C\; ) = 1-\mu(A^C \cup B^C) \geq 1-(\mu(A^C)+\mu(B^C))$ and
\begin{align*}
&\PPP(\{ (\omega,s,t) \mid X_t(\omega) = C(\omega) , s \in \T \}^C) 
\\&=1-\PPP(\{ (\omega,s,t) \mid X_t(\omega) = C(\omega) , s \in \T \}) 
\\&=1-\PPP(\{ (\omega,t) \mid X_t(\omega) = C(\omega) \} \times \T) 
\\&=1-\PP(\{ (\omega,t) \mid X_t(\omega) = C(\omega) \} )
\\&=\PP(\{ (\omega,t) \mid X_t(\omega) \neq C(\omega) \} ).
\end{align*}

Otherwise, define $p(\omega, A) := \int \I_A(X_t(\omega)) \d P_T(t)$. Since $X_t$ is measurable and $P_T$ is a probability measure $p$ is a Markov kernel. Yet, since $\P[X_T = X_S] = 1$ for all $A \in \Sigma_\X$ we have $1 = \P[\I_A(X_T) = \I_A(X_S)] = \P[|\I_A(X_T) - \I_A(X_S)| = 0]$ and therefore for all $\varepsilon > 0$ we have $0 = \P[|\I_A(X_T) - \I_A(X_S)| \geq \varepsilon]$ thus $\E[|\I_A(X_T) - \I_A(X_S)|] \leq \varepsilon$ so 
\begin{align*}
0 &= \E[|\I_A(X_T) - \I_A(X_S)|] 
\\&= \iint \E[|\I_A(X_t) - \I_A(X_s)| \mid T=t,S=s] \d P_T(s) \d P_T(t)
\\&= \iint \E[|\I_A(X_t) - \I_A(X_s)|] \d P_T(s) \d P_T(t)
\\&= \E\left[ \iint |\I_A(X_t) - \I_A(X_s)| \d P_T(s) \d P_T(t) \right] 
\\&\geq \E\left[ \int \left|\I_A(X_t) - \int \I_A(X_s) \d P_T(s)\right| \d P_T(t) \right] 
\\&=  \iint \left|\I_A(X_t(\omega)) - p(\omega,A)\right| \d P_T(t) \d \P(\omega) 
\\&= \iint \left|\I_A(X_t(\omega)) - p(\omega,A)\right| \d \P(\omega) \d P_T(t) \geq 0
\\\Rightarrow \exists t_0 \in \T : 0 &= \int \left|\I_A(X_{t_0}(\omega)) - p(\omega,A)\right| \d \P(\omega)
,
\end{align*}
by Jensen's inequality and Fubini's theorem.
Thus, $p(\omega, \cdot) \in \{0,1\}$ $\P$-a.s. so there exists a $\P$ null set $N \subset \Omega$ such that $p(\omega,\cdot) \in \{0,1\}$ for all $\omega \in N^C$. Since $\X$ is countably generated $p(\omega,\cdot)$ is a Dirac-measure for every $\omega \in N^C$. Thus, there exists a set theoretical map $C : \Omega \to \X$ such that $p(\omega,A) = \delta_{C(\omega)}(A)$ for all $\omega \in N^C$ and $C(\omega) = x_0 \in \X$ for all $\omega \in N$. Yet, since $p(\cdot, A)$ is measurable the mapping $\omega \mapsto \I_A(C(\omega))$ must be measurable for all $A \in \Sigma_\X, \omega \in N^C$ and thus $\{\omega \mid C(\omega) \in A\} = \{\omega \mid \I_A(C(\omega)) = 1\} \in \A$ is measurable as also $\{x_0\} \in \Sigma_\X$. Hence, $C$ is a $\X$-value random variable and by construction $p(\omega,C(\omega)) = 1$ and thus $P_T[X_t(\omega) = C(\omega)] = 1$ $\P$-a.s.
}

Thus, equality is far too restrictive for dependent data. Yet, the definition of drift as ``a change of the distribution over time'' stems from the batch setup with independent data where we do not expect any change and are able to make arbitrarily precise measurements as discussed in \cref{sec:taxonomy}. If we expect some kind of temporal dynamics, like seasonality, it might be better to rephrase this as ``the distribution changes in an unexpected way'' where ``unexpected'' means ``in a way our model cannot cope with without retraining''. This is a more suitable description as shown in the following example:

\begin{example}
\label{emp:sin}
Let $\T, \X = \R$. For $\varepsilon \in [0,\pi]$ consider the family of stochastic processes $X_t^{(\varepsilon)}(\omega) = \sin(t+U^{(\varepsilon)}(\omega))$ with $U^{(\varepsilon)} \sim \mathcal{U}([0,2\pi-\varepsilon])$ uniformly distributed. 
Then $X_t^{(0)}$ is the only stationary process in this family. 
Yet, given two points $S,T \sim \mathcal{U}([0,2\pi])$ we can estimate $U$ from $X_T^{(\varepsilon)},X_S^{(\varepsilon)}$ correctly with probability 1.
\end{example}
\OOproof{
Observe that
$\E[ X_0^{(\varepsilon)} ] = \int_0^{2\pi - \varepsilon} \sin(x) \d x = 1-\cos(\varepsilon)$ and
$\E[ X_\varepsilon^{(\varepsilon)} ] = \int_0^{2\pi - \varepsilon} \sin(\varepsilon+x) \d x =\int_\varepsilon^{2\pi} \sin(x) \d x = \cos(\varepsilon)-1$. Stationarity in particular implies constant expectation. Therefore, $ \cos(\varepsilon)-1 = 1-\cos(\varepsilon) \Leftrightarrow \cos(\varepsilon) = 1 \Rightarrow \varepsilon = 0$ is the only solution that can admit a stationary process. Yet, for $\varepsilon = 0$ we have $\P[X_t^{(0)} \in A] = \int_0^{2\pi} \I_A[\sin(t+x)] \d x \overset{\textbf{!}}{=} \int_0^{2\pi} \I_A[\sin(x)] \d x = \P[X_0^{(0)} \in A]$ where $\textbf{!}$ holds since $\sin$ has period $2\pi$. Hence, $X_t^{(0)}$ is stationary.

Observe that for all $y \in [-1,1]$ we have $|\{x \mid \sin(x) = y\} \cap [0,2\pi)| \leq 2$. Thus, once we know $X_T^{(\varepsilon)}(\omega) = y$ there are at most two choices left for $U(\omega)$. Yet, the chance of $S \in \{T + \mathbb{Z}\pi\}$ is 0 and in every other case we can determine $U(\omega)$ uniquely. 
}

In the next section, we explore the idea of ``unexpected change'' in more detail, leading to our substitute for the notion of stationarity.

\subsection{Temporal Consistency\label{sec:def-consistency}}

In the last section, we discussed that stationarity is a sub-optimal choice to describe the notion of drift in a dependent setup, rather we should look for ``unexpected changes of a single stream''. We will now formalize 
the idea, referring to the new notion as \emph{consistency}.
Formally, we define a collection of functions that assign a loss to the paths. Consistency then refers to time-invariant losses.

\begin{definition}
    Let $P_T$ be a probability measure on $\T$. Let $\F$ be a collection of measurable functions from $\T \times \X$ to $\R \cup \{\infty\}$. A path $x_t : \T \to \X$ is \emph{consistent with respect to $\F$} iff $\inf_{f^* \in \F} \int f^*(x_t,t) \d P_T(t) < \infty$ and equality holds for 
    \begin{align*}
        \inf_{f^*_1,\dots,f^*_n \in \F \atop W_1,\dots,W_n \in \mathcal{W}}  \sum_{i = 1}^n\int_{W_i} f^*_i(x_t,t) \d P_T(t) \leq \inf_{f^* \in \F} \int f^*(x_t,t) \d P_T(t)
    \end{align*}
    where $\mathcal{W}$ are all finite disjoint coverings of $\T$ by time windows.
\end{definition}

Consistency can be used to check whether $x_t$ belongs to a predefined function class $\mathcal{C}$ but also generalizes the notion of model drift as considered by \cite{hinder2023hardness} and is closely connected to $\H$-model drift~\cite{ida2023}:
\begin{lemma}
\label{lem:constypes}
Let $\X \subset \R^d$ and $\T = [0,1]$. Let $\mathcal{C}$ be a colleciton of measurable functions from $\T$ to $\R^d$ and $\H$ a hypothesis class with loss $\ell$ on $\X$. It holds:
\begin{enumerate}
    \item Consider $\F = \{ f(x,t) = \1[x \neq g(t)] \cdot \infty \mid g \in \mathcal{C} \}$ then $x_t$ is $\F$-consistent if and only if $x_t \in \mathcal{C}$ in the $L^0$-sense.
    \item Consider $\F = \{ f(x,t) = \Vert x-g(t) \Vert_2 \mid g \in \mathcal{C} \}$ and assume that the functions in $\mathcal{C}$ are continuous, and that for all $t$ up to a $P_T$-null set the closure of the images of $t$ under $\mathcal{C}$ contain $\X$, i.e., for all $t \in \T \setminus N$ and $x \in \X, \varepsilon > 0$ there exists a $g \in \mathcal{C}$ such that $\Vert g(t) - x\Vert < \varepsilon$, then $x_t$ can be approximated in $L^{p},\; 1 \leq p < \infty$ by cadlag function piecewise contained in $\mathcal{C}$. In particular, $x_t$ is $\F$-consistent if and only if $x_t \in \overline{\mathcal{C}}$ where the closure is taken in the $L^p$-sense. 
    \label{lem:constypes:l2}
    \item Consider $\F = \{f(x,t) = \ell(h,x) \mid h \in \H\}$ then $x_t$ is not $\F$-consistent implies that the drift process $(\delta_{x_t}, P_T)$ has $\H$-model drift~\cite{ida2023}.
\end{enumerate}
\end{lemma}
\OOproof{
1) if $f(x_t,t) = \infty$ on a $P_T$ non-null set then $x_t$ not consistent, otherwise $x_t = g(t)$ $P_T$-a.s. 

2) %
We show that we can approximate $x_t$ by functions piecewise contained in $\mathcal{C}$. To do so we first show that we can approximate $x_t$ by piecewise constant, cadlag functions with values in $\X$ and then that a constant function can be approximated by a cadlag function that is picewise contained in $\mathcal{C}$.

Let $\varepsilon > 0$ and
$y_0 \in \X$ be some point such that $\Vert x_t-y_0 \Vert_{L^{p,2}} \leq C < \infty$ for some $C \geq 1$. Such a point exists as we can approximate $x_t$ by a continuous function with values in all of $\R^d$ which is bounded since $\T$ is compact, this implies the existence by use of the triangle inequality. Since $\R^d$ is polish, $P_T \circ x_t^{-1}$ is a tight measure and thus there exists a compact set $K \subset \R^d$ such that $(P_T \circ x_t^{-1})(K^C) < \frac{\varepsilon}{4 C}$ and by Heine-Borel we may assume $K = \overline{B}_r(y_0)$ for some $r < \infty$. Define $S_0 = K^C$ and $$S_{-1} = \bigcup_{x \in K \::\: \inf \{ \Vert x-y \Vert_2 \mid y \in \X\} > \varepsilon/8 } B_{\varepsilon/8}(x).$$ The set $\{ B_{\varepsilon/4}(y) \mid y \in \X \} \cup \{S_{-1}\}$ is an open cover of $K$ and thus admits a finite sub-cover $S_{-1},S_1,\dots,S_n$ with associated points $y_1,\dots,y_n$. Let $W_0,W_1,\dots,W_n$ the pre-images of $S_0,S_1,\dots,S_n$ under $x_t$ (notice that the pre-image of $S_{-1}$ is empty). Since $\T$ is polish, $P_T$ is regular and thus there exist open sets $O_0,O_1,\dots,O_n$ such that $W_i \subset O_i$ and $P_T(O_i \setminus W_i) < \frac{\varepsilon}{4(2r+C)(n+1)}$ then denoting $p_0=1,p_1=P_T(W_1),\dots,p_n=P_T(W_n)$ we have
\begin{align*}
    \Vert (x_t-y_i)\I_{O_i} \Vert_{L^{p,2}} 
      &= \underbrace{\Vert (x_t-y_i)\I_{W_i} \Vert_{L^{p,2}}}_{\overset{!^1}{\leq} \varepsilon/4 \cdot p_i} + \Vert (x_t-y_i)\I_{O_i \setminus W_i} \Vert_{L^{p,2}}
    \\&\leq \frac{\varepsilon}{4}p_i + \underbrace{\Vert (x_t-y_0)\I_{O_i \setminus W_i} \Vert_{L^{p,2}}}_{\overset{!^2}{\leq} C \frac{\varepsilon}{4(2r+C)(n+1)}} + \underbrace{\Vert y_0-y_i \Vert_2 P_T(O_i \setminus W_i) }_{\leq 2r \frac{\varepsilon}{4(2r+C)(n+1)} }
    \\&\leq \frac{\varepsilon}{4} p_i + \frac{\varepsilon (2r+C)}{4 (2r+C) (n+1)}
    \\\Rightarrow \sum_{i = 0}^{n} \Vert (x_t-y_i)\I_{O_i} \Vert_{L^{p,2}}  &= \frac{\varepsilon}{4} \Big( p_0 + \underbrace{\sum_{i = 1}^n p_i}_{\leq 1} \Big) + \frac{\varepsilon}{4}
    \leq \frac{3\varepsilon}{4} < \varepsilon,
\end{align*}
where $!^1$ holds for $i = 0$ because $\Vert (x_t-y_0)\I_{W_0} \Vert_{L^{p,2}} \leq \Vert x_t-y_0 \Vert_{L^{p,2}} P_T(W_0) = \Vert x_t-y_0 \Vert_{L^{p,2}} (P_T\circ x_t^{-1})(K^C) \leq C \frac{\varepsilon}{4C}$ where the first inequality is Hölder's inequality with $q=\infty$ applied to the indicator function, for $i = 1,\dots,n$ we have $W_i = x_t^{-1}(B_{\varepsilon/4}(y_i))$ so $\Vert x_t - y_i \Vert < \varepsilon/4$, and $!^2$ holds by a similar application of Hölder's inequality. Thus, any piecewise constant function $g$ with $g^{-1}(y_i) \subset O_i$ is a good approximation of $x_t$. To construct $g$ consider the open covering $\{ B_{l/2}(t) \mid t \in O_i, B_l(t) \subset O_i, \; i = 0,\dots,n \}$ of $\T$. This is a cover indeed, as every $t \in \T$ is contained in some $W_i \subset O_i$. Choose a finite sub-cover $B_1,\dots,B_m$ and points $0 = t_1 < \dots < t_o < t_{o+1} = 1$ such that for all $k$ there is a $j$ and thus a $i$ such that $(t_k,t_{k+1}) \subset B_j \subset O_i$. By construction we have that if $(t_k,t_{k+1}) \subset O_i$ then also $[t_k,t_{k+1}),[t_k,t_{k+1}] \subset O_i$ and we can hence obtain an approximation of $x_t$ by a picewise constant, cadlag function. Hence, if we can approximate constant functions by functions in $\mathcal{C}$ we are done.

Let $\varepsilon > 0$ and assume
$x_t = x$ is constant. Denote by $B_r(t) = (t-r,t+r) \subset \R$ and extend all function in $\mathcal{C}$ by a constant onto all of $\T \subset \R$.
Consider $U = \{ B_{r/2}(t) \mid \exists g \in \mathcal{C} : \sup_{s \in B_r(t)} \Vert x-g(s) \Vert_2 < \varepsilon/3 \}$. Notice that every $t \in \T \setminus N$ is contained in some ball in $U$ as by assumption there exists a $g \in \mathcal{C}$ such that $\Vert g(t) - x \Vert < \varepsilon/6$ and thus $B_r(t) \subset g^{-1}[ B_{\varepsilon/6}(x) ]$ for some $r > 0$ as it is the pre-image of an open set under a continuous function. 

Fixate a function $g_0 \in \mathcal{C}$. Since $\T$ is compact $g_0$ is bounded and thus $\Vert x-g_0 \Vert_\infty < \infty$. Furthermore, as $\T$ is polish, $P_T$ is regular and thus there exists an open set $O$ such that $P_T(O \setminus N) < \frac{\varepsilon}{3\Vert x - g_0 \Vert_\infty}$. Define $V = \{ B_{\inf \{ \Vert s-t \Vert_2 / 2 \mid s \in O^C\} }(t) \mid t \in O\}$. Notice that every $t \in O$ is contained in some ball in $V$ so in particular every $t \in N$ is contained in some ball in $V$. 

Hence, $U \cup V$ is an open cover of $\T$ which admits a finite sub-cover $U_0 \cup V_0$ as $\T$ is compact. Thus, there exists a sequence $0 = t_0 < t_1 < \dots < t_n = 1$ such that $(t_i,t_{i+1}) \subset B$ with $B \in V_0 \cup U_0$. For every $(t_i,t_{i+1}) \in B$ with $B \in V_0$ we have that $[t_i,t_{i+1}] \subset O$ as we halved the radii, so by choosing $g_0$ for all those the error is still below $\Vert g_0-x \Vert_\infty P_T(O) = \Vert g_0-x \Vert_\infty (P_T(O \setminus N) + P_T(N)) \leq \Vert g_0-x \Vert_\infty (\frac{\varepsilon}{3 \Vert g_0-x \Vert_\infty} + P_T(N)) < \varepsilon/3$. Similarly, for every $(t_i,t_{i+1}) \in B$ with $B \in U_0$ there is a function $g$ associated with it such that $\sup_{s \in [t_i,t_{i+1}]} \Vert x-g(s) \Vert_2 < \varepsilon/3$ as we halved the radii. Combining all the pieces we obtain the desired function.

3) 
Assuming the left-hand side is strictly smaller than the right-hand side. Then there has to exist a $n$ and $f_1^*,\dots,f_n^*$ for which strict inequality already holds. Let us assume that $n$ is minimal, then all $f_i^*$ have to be different as otherwise we could use $n-1$ which is smaller. On the other hand, it holds $n \geq 2$ as otherwise both sides have to be equal. Thus there have to exist at least two disjoint windows $W_i,W_j$ and $f_i^*,f_j^*$ with $\int f_i^*(x_t,t) \d P_T(t \mid W_i) < \int f_j^*(x_t,t) \d P_T(t \mid W_i)$ and analogous for $j$. By definition of $\F$ and minimality of $n$ this implies $\H$-model drift using that $\int f^*_i(x_t,t) \d P_T(t\mid W_i) = \iint f^*_i(x,t) \d \delta_{x_t}(x) \d P_T(t\mid W_i) = \iint \ell(h,x) \d \delta_{x_t}(x) \d P_T(t\mid W_i) = \mathcal{L}_{(\delta_{x_t})_{W_i}}(h)$.
}
Thus, we define which functions are consistent. In the next section, we derive an algorithm that checks for consistency for just this setup.

\subsection{Measuring Consistency of a Noisy Stochastic Processes\label{sec:method}}
To apply consistency, we show how to check whether a stochastic process is essentially contained in a predefined function class $\mathcal{C}$ comparable to \cref{lem:constypes}.\ref{lem:constypes:l2}.
We consider a noisy (measured) version of the function $f$, i.e., $X_t = f(t) + \varepsilon_t$ where $\varepsilon_t$ is an independent noise process, and we want to check whether $f \in \mathcal{C}$. To do so we use a newly designed loss that can compensate for the noise. 
We assume that $\mathcal{C}$ specifies global phenomena, e.g., linear, polynomial, or exponential trends, periodicity, etc. Thus, $\mathcal{C}$ contains no local phenomena. 

An algorithmic solution is then given by the following: We fit a model $\hat{f} \in \mathcal{C}$ to $X_t$ and then check whether $f-\hat{f}$ does contain any structure using a local model like $k$-NN. If this is not the case, i.e., the local model is not better than random chance, then $f \in \mathcal{C}$. 

In case the paths are equidistantly sampled, the $k$-NN simplifies to the following loss which has the desired properties.
\begin{theorem}
\label{thm:lmse}
    Let $\T = [0,1]$ with sampling time points $t_i^{(n)} = \sfrac{i}{n}$. 
    Let $k : \N \to \N$ such that $k(n) \leq n$ and odd, denote by $k_0(n) = (k(n)-1)/2$. 
    Let $f,\hat{f} : \T \to \X$ be measurable maps and
    $X_t = f(t)+\varepsilon_t$ with $\varepsilon_t \indp \varepsilon_s \forall t \neq s, \E[\varepsilon_t] = 0$ and denote by $X_t^{(n)}$ the subsample according to $t_i^{(n)}$.
    We define the \emph{$k$-local MSE} as:
    \begin{align*}
        \lMSE_k(\hat{f} \mid X_t^{(n)}) = \frac{1}{n-k(n)} \sum_{i = k_0(n)}^{n-k_0(n)} \left(\frac{1}{k(n)} \sum_{j = -k_0(n)}^{k_0(n)} X^{(n)}_{t^{(n)}_{i+j}}-\hat{f}(t^{(n)}_{i+j})  \right)^2.
    \end{align*}
    If $|f(t)|,|\hat{f}(t)| < C$, $k(n) < n$, and $\Var(\varepsilon_t),\E[\varepsilon_t^4]$ are bounded, then it holds 
    \begin{align*}
        \left| \E[\lMSE_k(\hat{f} \mid X_t^{(n)})] - \MSE(\hat{f} \mid f) \right| \leq \max\left\{\frac{\sum_{i = 1}^n \Var\left(\varepsilon_{t_i^{(n)}}\right)}{k(n)(n-k(n))}, \frac{4C^2 \cdot k(n)}{n-k(n)} \right\}.
    \end{align*}
    And, if $k(n) \to \infty$ and $k(n)/n \to 0$ as $n \to \infty$ then the $k$-local MSE of the empirical process converges in probability to the MSE of the true signal as $n \to \infty$. 
\end{theorem}
\OOproof{
We have
\begin{align*}
    &\frac{1}{n-k(n)} \sum_{i = k_0(n)}^{n-k_0(n)} \left(\frac{1}{k(n)} \sum_{j = -k_0(n)}^{k_0(n)} X^{(n)}_{t^{(n)}_{i+j}}-\hat{f}(t^{(n)}_{i+j})  \right)^2
    \\&= \frac{1}{n-k(n)} \sum_{i = k_0(n)}^{n-k_0(n)} \left(\left(\frac{1}{k(n)} \sum_{j = -k_0(n)}^{k_0(n)} \varepsilon_{t^{(n)}_{i+j}}\right)+\left(\frac{1}{k(n)} \sum_{j = -k_0(n)}^{k_0(n)} f(t^{(n)}_{i+j})-\hat{f}(t^{(n)}_{i+j}) \right)\right)^2
    \\&=\underbrace{\frac{1}{n-k(n)} \sum_{i = k_0(n)}^{n-k_0(n)} \left(\frac{1}{k(n)} \sum_{j = -k_0(n)}^{k_0(n)} \varepsilon_{t^{(n)}_{i+j}}  \right)^2}_{(1)}
    \\&+2\underbrace{\frac{1}{n-k(n)} \sum_{i = k_0(n)}^{n-k_0(n)} \left(\frac{1}{k(n)} \sum_{j = -k_0(n)}^{k_0(n)} \varepsilon_{t^{(n)}_{i+j}}\right) \cdot \left(\frac{1}{k(n)} \sum_{j = -k_0(n)}^{k_0(n)} f(t^{(n)}_{i+j})-\hat{f}(t^{(n)}_{i+j}) \right)}_{(2)}
    \\&+ \underbrace{\frac{1}{n-k(n)} \sum_{i = k_0(n)}^{n-k_0(n)} \left(\frac{1}{k(n)} \sum_{j = -k_0(n)}^{k_0(n)} f(t^{(n)}_{i+j})-\hat{f}(t^{(n)}_{i+j})  \right)^2}_{(3)}.
\end{align*}
Since $\E[\varepsilon_t^{(n)}] = 0$ the second term vanishes and the first term equals 
\begin{align*}
(1) = \frac{1}{k(n)^2(n-k(n))} \sum_{i = k_0(n)}^{n-k_0(n)} \sum_{j = -k_0(n)}^{k_0(n)} \Var\left(\varepsilon_{t^{(n)}_{i+j}}\right)    
\end{align*}
using that all $\varepsilon_{t_i^{(n)}}$ are independent. By reordering we see that $\Var\left(\varepsilon_{t^{(n)}_{i}}\right)$ occurs 
at most $k(n)$ times, thus, the first term is bounded by 
\begin{align*}
    (1) \leq \frac{1}{k(n)(n-k(n))} \sum_{i = 1}^{n} \Var\left( \varepsilon_{t^{(n)}_{i}}  \right).
\end{align*}
Similarly, by Jensen's inequality the third term is bounded by the MSE using the same reordering where we have to add at most $k(n)^2-1$ terms in case $k(n)=n$ each bounded by $4C^2$. Hence, we have
\begin{align*}
    0 \leq \MSE(\hat{f} \mid f)-(3) \leq \frac{4C^2 \cdot k(n)}{n-k(n)}.
\end{align*}
Together with the other bound we thus obtain
\begin{align*}
    -\frac{4C^2 \cdot k(n)}{n-k(n)} \leq \underbrace{(3)-\MSE(\hat{f} \mid f)}_{\leq 0} + (1) \leq \frac{1}{k(n)(n-k(n))} \sum_{i = 1}^{n} \Var\left( \varepsilon_{t^{(n)}_{i}}  \right)
\end{align*}
and we arrive at the statement by taking absolute values. 

Under the assumptions on $k(n)$ we have that $\frac{k(n)}{n-k(n)} = \frac{k(n)}{n}\frac{1}{\left(1-\frac{k(n)}{n}\right)}$, as $\frac{k(n)}{n} \to 0$ the first term goes to 0 and the second to 1 so the product converges to 0 and therefore $\frac{4C^2 \cdot k(n)}{n-k(n)} \to 0$. Furthermore, denoting by $V$ an upper bound of all $\Var(\varepsilon_t)$ and using $k(n)(n-k(n)) = k(n)n\left(1-\frac{k(n)}{n}\right)$ we can bound
\begin{align*}
    \frac{\sum_{i = 1}^{n} \Var\left( \varepsilon_{t^{(n)}_{i}}  \right)}{k(n)(n-k(n))} 
    &\leq \frac{n V}{k(n)n\left(1-\frac{k(n)}{n}\right)} \\&= \underbrace{\frac{V}{k(n)}}_{\to 0} \cdot \underbrace{\frac{1}{1-\frac{k(n)}{n}}}_{\to 1} \to 0
\end{align*}
so the expectation of the local $k$-MSE converges to the MSE of the original signal as $n \to \infty$.

To show that the empirical local $k$-MSE converges to its expectation, we show that the variance tends to 0 as $k(n) \to \infty$. To do so, first observe that if $Y_1,...,Y_n$ are random variables with $\Var(Y_i) < V'$, then 
\begin{align*}
    \Var\left(\frac{1}{n-k(n)} \sum_{i = k_0(n)}^{n-k_0(n)} Y_i\right) 
    &= \frac{1}{(n-k(n))^2} \sum_{i = k_0(n)}^{n-k_0(n)}\sum_{j = k_0(n)}^{n-k_0(n)} \text{Cov}(Y_i,Y_j)
    \\&\leq \frac{1}{(n-k(n))^2} \sum_{i = k_0(n)}^{n-k_0(n)}\sum_{j = k_0(n)}^{n-k_0(n)} \sqrt{\Var(Y_i)}\sqrt{\Var(Y_j)}
    \\&= \left(\frac{1}{n-k(n)} \sum_{i = k_0(n)}^{n-k_0(n)} \sqrt{\Var(Y_i)} \right)^2
    \\&\leq \left(\frac{1}{n-k(n)} \sum_{i = k_0(n)}^{n-k_0(n)} \sqrt{V'} \right)^2
    \\&= (\sqrt{V'})^2 = V'.
\end{align*}
Thus, by identifying $Y_i = \left(\frac{1}{k(n)} \sum_{j = -k_0(n)}^{k_0(n)} X^{(n)}_{t^{(n)}_{i+j}}-\hat{f}(t^{(n)}_{i+j})  \right)^2$ we see that it remains to show that $\Var(Y_i) \to 0$ as $k(n) \to \infty$.

By writing
\begin{align*}
    Y_i &= \left(\frac{1}{k(n)} \sum_{j = -k_0(n)}^{k_0(n)} \varepsilon_{t_{i+j}^{(n)}} \right)^2 +\left(\frac{1}{k(n)} \sum_{j = -k_0(n)}^{k_0(n)} f(t^{(n)}_{i+j})-\hat{f}(t^{(n)}_{i+j}) \right)^2 \\&+ 2\left(\frac{1}{k(n)} \sum_{j = -k_0(n)}^{k_0(n)} \varepsilon_{t_{i+j}^{(n)}} \right)\left(\frac{1}{k(n)} \sum_{j = -k_0(n)}^{k_0(n)} f(t^{(n)}_{i+j})-\hat{f}(t^{(n)}_{i+j}) \right) 
\end{align*}
we see by using $\text{Cov}(A,B) \leq \sqrt{\Var(A)\Var(B)}$ that is suffices to consider
\begin{align*}
    &\Var\left( \left(\frac{1}{k(n)} \sum_{j = -k_0(n)}^{k_0(n)} \varepsilon_{t_{i+j}^{(n)}} \right)\left(\frac{1}{k(n)} \sum_{j = -k_0(n)}^{k_0(n)} f(t^{(n)}_{i+j})-\hat{f}(t^{(n)}_{i+j}) \right) \right) 
    \\&\leq 4C^2 \Var\left( \frac{1}{k(n)} \sum_{j = -k_0(n)}^{k_0(n)} \varepsilon_{t_{i+j}^{(n)}} \right)
    \\&\leq \frac{4C^2 V}{k(n)}
\end{align*}
and with $W$ an upper bound of $\E[\varepsilon_t^4]$ we have
\begin{align*}
    \Var\left(\left(\frac{1}{k(n)} \sum_{j = -k_0(n)}^{k_0(n)} \varepsilon_{t_{i+j}^{(n)}} \right)^2\right) 
    &\leq \E\left[\left(\frac{1}{k(n)} \sum_{j = -k_0(n)}^{k_0(n)} \varepsilon_{t_{i+j}^{(n)}} \right)^4\right]
    \\&\overset{!}{=} \frac{1}{k(n)^4} \sum_{j,j' = -k_0(n)}^{k_0(n)} \E\left[ \varepsilon_{t_{i+j}^{(n)}}^2\varepsilon_{t_{i+j'}^{(n)}}^2 \right]
    \\&=\frac{(W-V) k(n) + V k(n)^2}{k(n)^4} \leq \frac{\max\{W,V\}}{k(n)^2},
\end{align*}
where $!$ holds because every term $\E[\varepsilon_{t_1}\varepsilon_{t_2}\varepsilon_{t_3}\varepsilon_s] = \E[\varepsilon_{t_1}\varepsilon_{t_2}\varepsilon_{t_3}]\E[\varepsilon_s] = 0$ for all $t_1,t_2,t_3$ and any$s \not\in \{t_1,t_2,t_3\}$.
So $\Var(Y_i) \leq \frac{\max\{W,V\}}{k(n)^2} + \frac{4C^2 V}{k(n)} + \frac{2C\sqrt{\max\{W,V\}V}}{\sqrt{k^3(n)}} \to 0$.
}

The $k$-local MSE is essentially the usual MSE except that smoothing is applied to the training data and the model. Indeed, for linear models with non-linear preprocessing, this allows for efficient training. Yet, in our experiments, training with the usual MSE usually leads to better results and the local MSE is only used for testing.

In the next section, we empirically evaluate the different methods to probe consistency. We also check the validity of the arguments presented in \cref{sec:stationarity} using numerical evidence.

\section{Numerical Evaluation\label{sec:exp}}
We perform numerical evaluations on artificial data to confirm the presented theory experimentally.\footnote{The experimental code as well as all datasets can be found at \url{https://github.com/FabianHinder/Remark-on-Dependent-Drift}.}

\subsection{Testing Stationarity}

We start by verifying that stream learning algorithms indeed operate path-wise and thus are not well described using the notion of stationary. To do so we generate paths according to the distribution in \cref{emp:jump} and \cref{emp:sin} in a stationary and non-stationary case. We then apply the Kernel Change-point Detector~\cite{KCpD} (KCpD) to the resulting paths. In the case of \cref{emp:jump} the detector determines the value of $C$ (the moment of the jump) with less than 2 samples offset each time, independent of whether the distribution was stationary or not. Similarly, for \cref{emp:sin} every point is considered a change point. Furthermore, using the sub-sampling process described in \cref{thm:sampling} the detector alerts a drift if and only if the initial distribution is not stationary, as predicted by the theorem. 
We also applied the Augmented Dickey-Fuller~\cite{ADF} (ADF), and the Kwiatkowski-Phillips-Schmidt-Shin~\cite{KPSS} (KPSS) test which are used to test for stationarity in time series to the data. The resulting $p$-values do not differ for the stationary and non-stationary cases for both tests. 
We thus conclude that the methods do not check for the stationarity of the underlying distribution or rather fail to do so but rather for properties of the single path.

\begin{figure}[!t]
    \centering
    \includegraphics[width=\textwidth]{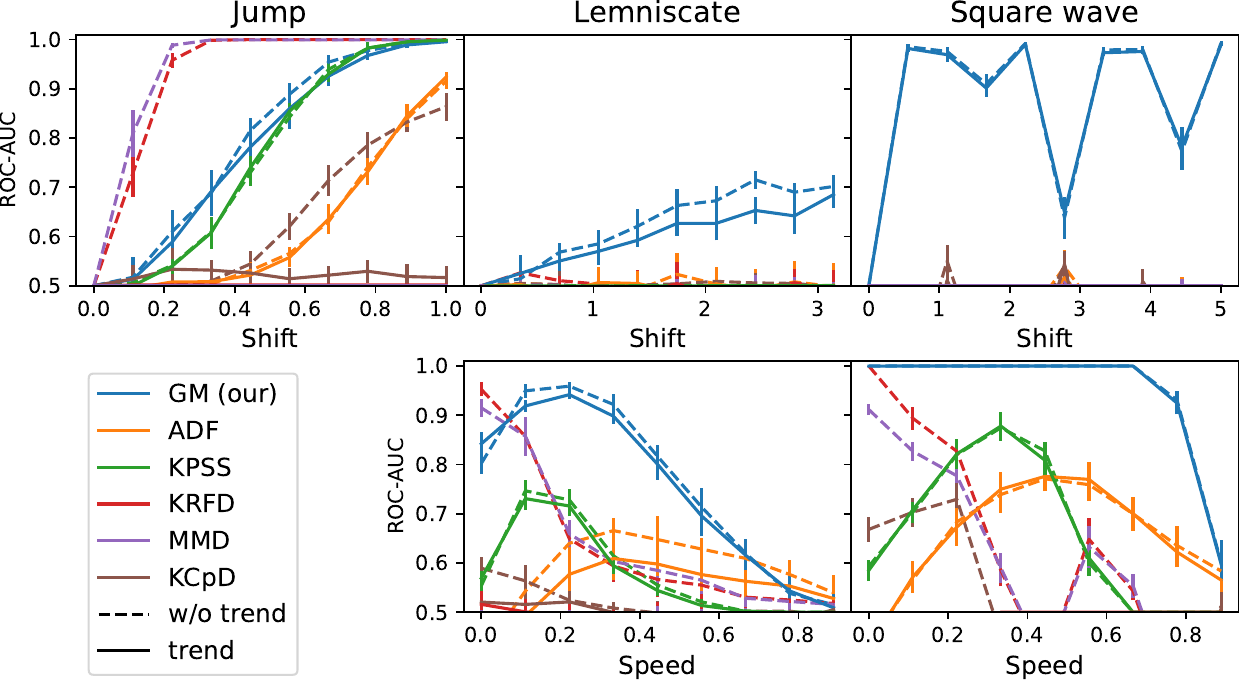}
    \caption{Mean (and standard deviation) of results over $10\times100$ runs. $x$-axis shows time shift/slow down and $y$-axis ROC-AUC-Score.}
    \label{fig:results}
\end{figure}

\subsection{Evaluation of Method}
To evaluate how well existing methods can deal with consistency we consider the following synthetic datasets: ``Jump'' similar to \cref{emp:jump}, ``Lemniscate'' $f(t) = C_1 \cos(t) + C_2 \sin(2t)/2$, and ``Square  wave'' where $f$ is given by a square wave. In all cases, we added normal distributed noise and considered a version with an additive linear trend (which is not considered as drift). Drift is induced in the middle of the path. In the case of Jump a jump in the mean value is introduced, for Lemniscate and Square wave we either added a shift in time, i.e., $f(t)$ for $t<0.5$, $f(t+\Delta), \Delta > 0$ for $t\geq 0.5$, or slowed down time, i.e., $f(t)$ for $t<0.5$, $f(ct), c \in [0,1)$ for $t\geq 0.5$. All with different intensities. We considered ADF, KPSS, and KCpD as well as MMD~\cite{MMD} and KFRD~\cite{KFRD} as methods from the literature, where the latter two are provided with the correct change point candidate. ADF and KPSS performed trend corrections. For KCpD we used the number of found change points, for all other methods the returned $p$-value.\footnote{We also considered auto-regressive models based on Ridge and $k$-NN regression but found them to only work on the Jump dataset.}
As suggested by \cref{thm:lmse}, we use the lMSE as an indicator for non-consistency using a linear model with (trig-)polynomial (up to degree 15) as preprocessing (GM). For every setup, we performed 1{,}000 independent runs and evaluated the obtained scores using the ROC-AUC to measure how well the methods separate the drifting and non-drifting paths~\cite{hinder2022suitability}. The results are presented in \cref{fig:results}.

As can be seen, the classical drift detectors KCpD, MMD, and KFRD only work for Jump without trend and in case of an extreme slow down, although KCpD and KFRD were explicitly designed for time series data \cite{KCpD,KFRD}. The other methods are less affected by the trend. ADF and KPSS perform quite well on Jump and in the cases of slow down. However, none of the methods from the literature can deal with a time shift on any dataset. Our method performs quite well on all datasets. In particular, it is the only one that can deal with time shifts.

\section{Conclusion\label{sec:concl}}
In this work, we argued that drift and stationarity do not align in many setups realistic for online learning and monitoring of dependent data. Next to providing theoretical counter-examples, we ran a numerical evaluation confirming our considerations. Besides, we proposed the concept of consistency which is more suitable to the tasks at hand. 

Yet, it is still unclear how problematic the discrepancy between stationarity and drift is. Also, besides the notion of (strict) stationarity we studied in this paper, the notion of wide-sens stationarity might offer a well-suited choice for some cases. An in-depth investigation of this seems interesting.
Furthermore, so far we have only considered the notion of consistency for stochastic processes. A description of the phenomenon of dependent data streams by means of ``dependent drift process'' and a suited generalization of the notion of consistency are still outstanding. 

Furthermore, the notion of consistency is very similar to the notion of model drift studied in~\cite{hinder2023hardness} who show that such approaches are not suited for monitoring tasks in the independent setting. Though we believe that this does not pose a problem as the proof is heavily founded on the independence assumption, considering this problem in detail appears to be very relevant.  

\bibliographystyle{splncs04}
\bibliography{bib}

\end{document}